\documentclass[a4paper,twoside]{article}

\usepackage{booktabs}
\usepackage[table]{xcolor}
\usepackage[nolist, nohyperlinks]{acronym}
\usepackage{multirow}
\usepackage{epsfig}
\usepackage{subcaption}
\usepackage{calc}
\usepackage{amssymb}
\usepackage{amstext}
\usepackage{amsmath}
\usepackage{amsthm}
\usepackage{multicol}
\usepackage{pslatex}
\usepackage{apalike}
\usepackage{SCITEPRESS}

\definecolor{sc:indoor}{rgb}{0.82, 0.94, 0.75}
\definecolor{sc:driving}{rgb}{1.0, 0.94, 0.75}

\begin{document}

\title{SALT: A Semi-automatic Labeling Tool for RGB-D Video Sequences}

\author{\authorname{
Dennis Stumpf\sup{1}, 
Stephan Krau\ss\sup{1}, 
Gerd Reis\sup{1}, 
Oliver Wasenm\"uller\sup{2} and 
Didier Stricker\sup{1,3}}
\affiliation{\sup{1}German Research Center for Artificial Intelligence GmbH (DFKI), Germany}
\affiliation{\sup{2}Hochschule Mannheim, Germany}
\affiliation{\sup{3}University of Kaiserslautern, Germany}
\email{\{dennis.stumpf, stephan.krauss, gerd.reis, didier.stricker\}@dfki.de, o.wasenmueller@hs-mannheim.de}
}


\abstract{Large labeled data sets are one of the essential basics of modern deep learning techniques. Therefore, there is an increasing need for tools that allow to label large amounts of data as intuitively as possible. In this paper, we introduce SALT, a tool to semi-automatically annotate RGB-D video sequences to generate 3D bounding boxes for full six \ac{dof} object poses, as well as pixel-level instance segmentation masks for both RGB and depth. Besides bounding box propagation through various interpolation techniques, as well as algorithmically guided instance segmentation, our pipeline also provides built-in pre-processing functionalities to facilitate the data set creation process. By making full use of SALT, annotation time can be reduced by a factor of up to 33.95 for bounding box creation and 8.55 for RGB segmentation without compromising the quality of the automatically generated ground truth.}

\onecolumn \maketitle \normalsize \setcounter{footnote}{0} \vfill

\section{\uppercase{Introduction}}
\label{sec:introduction}
Generating ground truth data for RGB-D data sets is extremely time-consuming and expensive, as it involves annotations in both 2D and 3D. The 3D bounding boxes of the famous data sets KITTI \cite{geiger2013kitti} and SUN RGB-D \cite{song2015sun-rgbd}, for example, are created manually for every individual object in every frame or scene. Furthermore, as such tasks are often outsourced to workers with limited technical background, the annotation quality can vary highly. KITTI, for example, contains frames with missing or incorrectly labeled boxes. This raises the question, if the manual effort in the annotation pipeline for RGB-D data can be reduced without compromising quality.

\begin{figure}
	\centering
		\includegraphics[width=\linewidth]{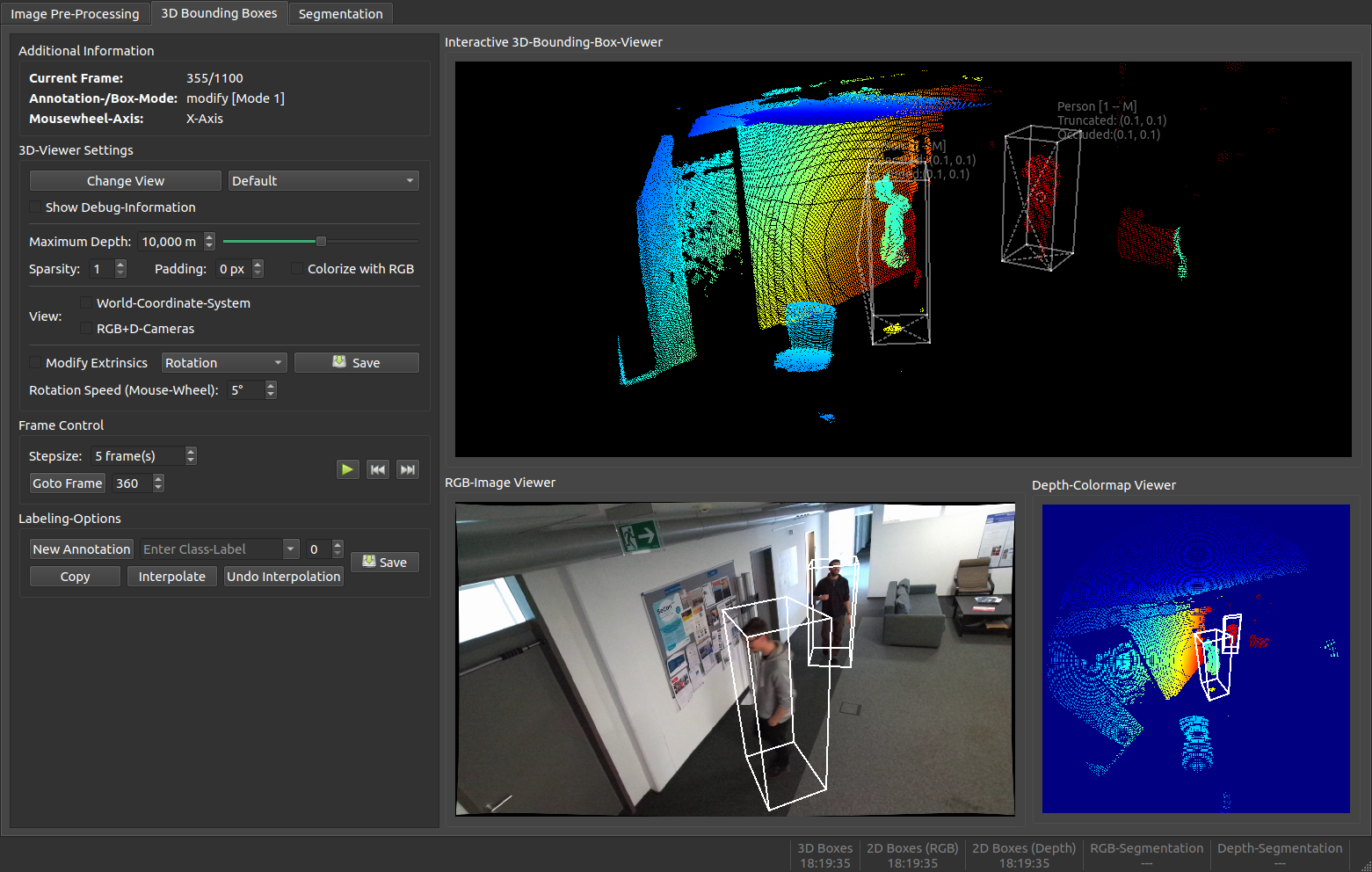}
		\caption{We introduce SALT, a tool to semi-automatically annotate 3D bounding boxes and pixel-level instance segmentation in RGB-D video sequences. This figure depicts a screenshot from the 3D bounding box module of the tool.}
		\label{fig:tool_screenshot}
\end{figure}

As it can be seen in Section \ref{sec:related_work}, this question opens up an active field of research where different approaches have already emerged. These approaches, however, all come with their individual limitations. They either require extensive pre-processing steps, are only applicable to static scenes containing rigid objects or limit the ground truth representation (e.g. 4 \ac{dof} object poses or segmentation for only one modality). Furthermore, many are designed for application scenarios involving an ego-motion of the camera.

To alleviate these and other limitations, we propose SALT (cf. Figure \ref{fig:tool_screenshot}), a simple, yet effective tool to generate bounding box and segmentation ground truth data for RGB-D video sequences in a semi-automatic fashion. The tool does not require any prior knowledge of the scene and is independent of any objects to be labeled. In summary, our contributions are:
\begin{enumerate}
	\item A pipeline to rapidly generate 3D bounding boxes for full 6 \ac{dof} object poses up to 33.95 times faster than a naive approach.
    \item Algorithmically guided generation of pixel-level instance segmentation masks in both 2D and 3D, with 2D bounding boxes as a by-product, speeding up annotation time by a factor of up to 8.55.
\end{enumerate}

\begin{table*}
    \centering
        \caption{Overview of state-of-the-art tools (upper half) and methods (lower half) for RGB-D data set creation. Listed are the detection and segmentation ground truth data representation, whether the tool provides built-in calibration functionalities (e.g. undistorting images or modifying extrinsic parameters) and if the functionalities of the tool are accessible through a \acf{gui}. Green rows indicate approaches built for annotating indoor scenarios (e.g. by using a Microsoft Kinect) and yellow ones for autonomous vehicles (i.e. LiDAR sensor, ego motion of cameras).}
        \label{tab:related_work}
		\begin{tabular}{c c c c c c}
	\toprule
	Method & Approach & Detection & Segmentation & Calibration & \acs{gui}\\
	\midrule
	\rowcolor{sc:indoor} Ours & interpolation + GrabCut & 6DoF, 2D & RGB, D & \checkmark & \checkmark \\	 
	\rowcolor{sc:indoor} \cite{marion2018labelfusion} & mesh + reconstruction & 6\acs{dof} & RGB, D & \checkmark & (\checkmark) \\
	\rowcolor{sc:indoor} \cite{suchi2019easylabel} & incremental scene building & 2D & D & - & - \\ 
	\rowcolor{sc:indoor} \cite{wong2015Smartannotator} & learning & 4\acs{dof} & RGB & - & \checkmark \\
	\rowcolor{sc:indoor} \cite{monica2017multilabel} & graph segmentation & - & D & - & \checkmark \\
	\rowcolor{sc:driving} \cite{3dbat} & interpolation & 4\acs{dof} & - & - & \checkmark \\
	\rowcolor{sc:driving} \cite{latte} & tracking + clustering & 4\acs{dof} & - & - & \checkmark \\
	\rowcolor{sc:driving} \cite{arief2020sane} & tracking + clustering & 4\acs{dof} & - & - & \checkmark \\
	\rowcolor{sc:driving} \cite{plachetka2018tubs} & tracking + optimization & 4\acs{dof}, 2D & RGB, D & \checkmark & \checkmark \\
	\rowcolor{sc:driving} \cite{lee2018leverage} & learning & 4\acs{dof} & D & - & \checkmark \\
	\rowcolor{sc:driving} \cite{wang2019apolloscape} & 3D-2D projection & - & RGB, D & - & \checkmark \\
	\rowcolor{sc:driving} \cite{yan2019lcas} & clustering & 3\acs{dof} & - & - & \checkmark \\
	\midrule
	\rowcolor{sc:indoor} \cite{hodan2017tless} & mesh + reconstruction & 6\acs{dof}, 2D & - && \\
	\rowcolor{sc:indoor} \cite{xiang2018posecnn} & mesh + reference frame & 6\acs{dof} & D && \\ 
	\rowcolor{sc:indoor} \cite{grenzdoerfer2020ycbm} & mesh + reference frame & 6\acs{dof}, 2D & D && \\ 
	\rowcolor{sc:driving} \cite{xie2016labeltransfer} & label transfer & - & RGB && \\
	\rowcolor{sc:driving} \cite{patil2019h3d} & interpolation & 4\acs{dof} & - && \\
	\rowcolor{sc:driving} \cite{dewan2016motiondetection} & Bayesian approach & 3\acs{dof} & D && \\
	\rowcolor{sc:driving} \cite{chang2019argoverse} & accumulation over time & 6\acs{dof} & - && \\
	\bottomrule
\end{tabular}

\end{table*}

\section{\uppercase{Related Work}}\label{sec:related_work}
Recent advances in the field of data set creation have shown that automating parts of the annotation process in order to minimize necessary human interaction can result in large-scale, high-quality RGB-D data sets in a short amount of time.

\cite{suchi2019easylabel} build a static scene by adding one object at a time, leveraging the change in depth for automatic pointwise segmentation. \cite{monica2017multilabel} use a nearest neighbor graph representation to segment a single point cloud based on individually labeled points. The quality of the generated ground truth of such approaches, however, highly depends on the depth quality and the texture of objects.

\cite{hodan2017tless} and \cite{marion2018labelfusion} reconstruct the scene, align pre-built meshes with the objects in the point cloud and then project back into the frames used for reconstruction. Similarly, \cite{xiang2018posecnn} and \cite{grenzdoerfer2020ycbm} align meshes with the point cloud of a reference-frame and then propagate the resulting object configuration to other frames. While mesh based approaches provide full 6 \ac{dof} object poses and segmentation masks, they require a pre-defined set of rigid objects and corresponding meshes beforehand. Thus, they are only capable of annotating known, static scenes.

\begin{figure*}
	\centering
		\includegraphics[width=\textwidth]{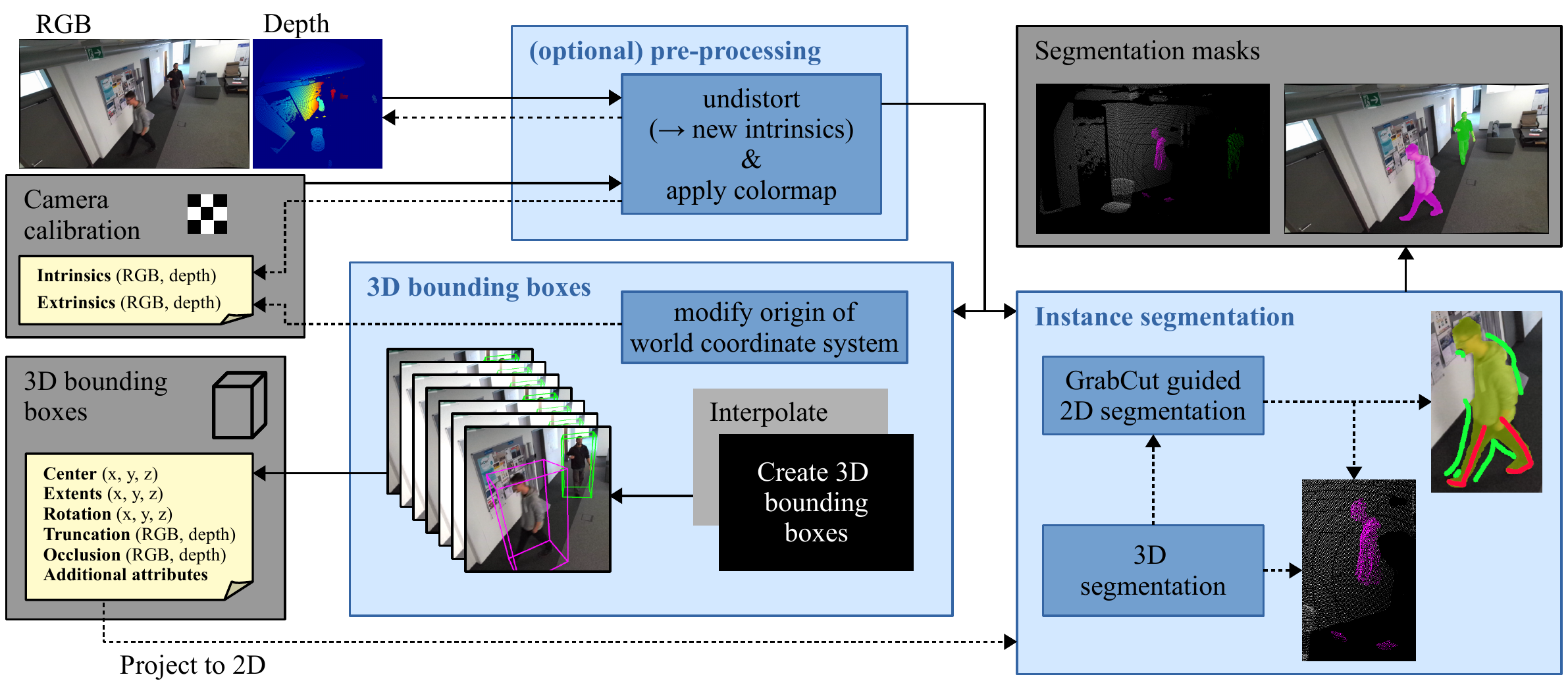}
		\caption{Overview of the pipeline of our proposed tool SALT. Input are the individual frames of an RGB-D sequence from a fixed viewpoint, as well as the corresponding camera calibration (intrinsics and extrinsic transformation between RGB and depth). The footage is either already undistorted, or can be undistorted using our tool. The 3D bounding box module can be used to not only generate 3D bounding boxes in a semi-automatic fashion, but also to modify the extrinsics of the RGB-D camera system w.r.t. the scene through simple drag and drop of the world coordinate system's origin. The instance segmentation module can segment depth maps and RGB images with the support of GrabCut \cite{rother2004grabcut}, or in the case of depth maps, by selecting the points directly from inside the point cloud. Already segmented depth maps can additionally aid the segmentation of the corresponding RGB image.}
		\label{fig:tool_overview}
\end{figure*}

\cite{wong2015Smartannotator} use actively trained models to predict the 3D structure of indoor scenes from segmented RGB-D images. \cite{lee2018leverage} leverage pre-trained \acp{cnn} to predict a 3D bounding box from a single click on an object's point in the point cloud. \cite{xie2016labeltransfer} use a \ac{crf} model to transfer a coarsely labeled 3D point cloud and geometric cues into 2D image space, resulting in densely segmented 2D images. Such learning based approaches, however, require annotated data beforehand, do not guarantee to scale to different data domains and are highly dependent on the used architectures.

\cite{latte}, \cite{arief2020sane} and \cite{yan2019lcas} infer 3D bounding boxes from the spatial extends of clustered points. The former two additionally provide \ac{cnn} assisted pre-segmentation and use tracking to propagate annotations into subsequent frames. \cite{dewan2016motiondetection} use a Bayesian approach and motion cues to automatically detect and track objects. \cite{plachetka2018tubs} track manually annotated 3D bounding boxes and corresponding polygons (for RGB segmentation). \cite{chang2019argoverse} accumulate manually selected points over time to automatically infer fixed size 3D bounding boxes. The drawback of these approaches is that the spatial extents of an object in a point cloud may not always represent the full spatial extents due to occlusion, sparse measurements or reflective textures. In such cases, tracking and clustering may yield inaccurate results. 

\cite{wang2019apolloscape} manually segment background and moving objects in 3D and 2D respectively and by combining both, they retrieve a fully segmented image. \cite{3dbat} and \cite{patil2019h3d}, similarly to our approach, annotate 3D bounding boxes for only a subset of frames, while interpolating the remaining ones. Both, however, only interpolate linearly, which may not accurately capture real dynamic motion.

\section{\uppercase{Method}}
\label{sec:method}
In this section, we describe the core features of SALT in greater detail. A graphical overview can be viewed in Figure \ref{fig:tool_overview}, which can be used as a guiding reference throughout this section.

\subsection{Pre-processing}
\label{subsec:preprocessing}
The tool internally uses undistorted RGB and depth frames (16 bit depth and 8 bit color mapped depth). The user has the option to either provide these images directly, or use the tool to generate them from the raw footage. In both cases, the appropriate intrinsic parameters of both cameras and the extrinsic stereo calibration (rotation $R$ and translation $t$ between the individual cameras) have to be provided.

\begin{figure}[ht]
	\centering
		\includegraphics[width=\linewidth]{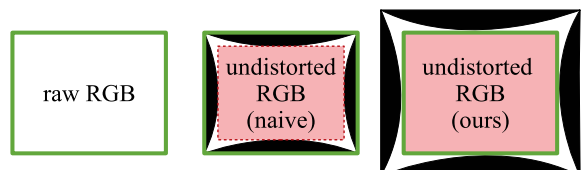}
		\caption{Simplified, schematic visualization of our undistortion pipeline for RGB images with radial distortion. The naive approach undistorts while retaining the original spatial dimension (green), resulting in loss of information through scaling. Our approach assures to keep the valid area (red) of roughly the original spatial dimension, retaining as much information (i.e. original pixels) as possible.}
		\label{fig:rgb_undistortion}
\end{figure}

Our pipeline undistorts in such a way that as many pixels of the raw input image as possible are retained. As it can be seen in Figure \ref{fig:rgb_undistortion}, undistorting an image while retaining the resolution and all pixels results in areas (indicated in black) that do not contain any pixel values. The naive approach would be to crop out the area containing only valid pixels (indicated in red) and scale it back to the original resolution. But this results in loss of information, as pixels outside of the valid area are dropped and pixels inside are already scaled down. Our pipeline, on the other hand, determines the ideal resolution at which one side of the valid area matches the original image resolution. This assures that during undistortion, as few pixels as possible are lost. Depth maps are processed in a similar fashion. The tool undistorts each pixel individually and projects them back to the original image space. Depending on the distortion model, the ideal coordinates of projected pixels could fall outside of the original resolutions' boundaries. To capture those pixels as well, the necessary resolution and focal lengths are computed and a new camera matrix is generated.

While we do not claim novelty for this undistortion process, we argue that it is necessary to generate a data set that is as flexible as possible. This means that, at a later point in time, if the images in the data set need to be undistorted differently, already generated segmentation masks lose as few pixels as possible when being re-distorted.

\subsection{3D Bounding Boxes}
\label{subsec:3d_bounding_boxes}
As depicted in Figure \ref{fig:tool_screenshot}, the user accesses the scene with an interactive 3D point cloud viewer. To identify objects in the scene and adjust parameters more easily, the boxes are projected and displayed in the corresponding RGB and depth frames in real time.

Let $\lbrace F_1, F_2, \ldots, F_N\rbrace $ be a sequence of $N$ RGB-D frames. When creating a new 3D bounding box $B_k$ for frame $F_k$, a user first assigns a unique interpolation ID and a class label. The process of drawing the box and adjusting its parameters involves simple drag-and-drop operations (translation and scaling) and the use of the mouse wheel (rotation) to reduce the cognitive workload.

Let the object contained in $B_k$ be visible for $M$ frames. In order to retrieve the boxes for all frames $B_{all} = \lbrace B_k, B_{k+1}, B_{k+2}, \ldots, B_{k+M}\rbrace $ with minimal effort, two core features are provided: \textit{copy} and \textit{interpolate}. Given the sequential nature of videos, copying the boxes into subsequent frames facilitates the annotation process, since only minor adjustments, if any, need to be made to copied boxes. This can then be used in conjunction with interpolation: The user only needs to fully annotate box $B_k$, copy it into every $m$-th frame (keyframe) and apply small adjustments, resulting in $B_{key} = \lbrace B_{k}, B_{k+m}, B_{k+2m}, \ldots, B_{k+M}\rbrace$. A simple interpolation $f_{\text{int}}(\cdot)$ then generates the box parameters for all intermediate frames. More formally:
\begin{equation}
    \label{eq:interpolation}
    B_{all} = f_{\text{int}}\left(B_{key}\right)
\end{equation}

The advantage of using a simple interpolation instead of, for example, a more complex tracking algorithm is its robustness against sparse or occluded measurements. Even objects without any points (i.e. visible only in the RGB frame) can be annotated.

Commonly, translation and scaling parameters are interpolated individually using simple linear interpolation (e.g. \cite{3dbat} and \cite{patil2019h3d}). We follow this approach for the scaling parameters (width, height and length), but implement a hybrid interpolation method for the translation parameters. More precisely, if the Euclidean distance $c_\Delta$ of the center coordinates is above a certain threshold $\epsilon$ for at least 4 consecutive keyframes, we apply cubic spline interpolation between those keyframes, otherwise linear interpolation is used:
\begin{equation}
    \label{eq:hybrid interpolation}
    f_\text{int}(\cdot) =
    \begin{cases}
      \text{cubic}  & \text{if $c_\Delta > \epsilon$ for $\geq$ 4 keyframes}\\
      \text{linear} & \text{otherwise}\\
    \end{cases}       
\end{equation}

Rotation angles (yaw, pitch and roll), on the other hand, cannot be interpolated individually. Given the Euler angles $\varphi$, a single 3D orientation is constructed from individual rotations around the $z$, $y$ and $x$ axis in a fixed order. For this reason, we apply \ac{slerp}, as introduced in \cite{shoemake1985slerp}. Instead of using three Euler angles, keypoint orientations are represented as quaternions. Thus an interpolation step involves moving between quaternions on a sphere around a fixed rotation axis with constant velocity, resulting in unambiguous rotational movement.

Furthermore, the tool automatically provides a truncation value $t$ by determining the cut-off area of bounding boxes at image boundaries when projected into the RGB and depth frames. Occlusion $o$, on the other hand, can be manually specified and will be propagated to interpolated boxes. By combining these values, a visibility score $v$ is computed as:
\begin{equation}
    \label{eq:visibility}
	v = (1 - t) \cdot (1 - o) \quad\quad t,o \in \left[0, 1\right]
\end{equation}

This value is used to colorize boxes depending on their visibility, in order to guide the user in visually understanding the scene and annotations more intuitively. Furthermore, the occlusion and truncation values can be used to categorize the difficulty of individual samples similarly to KITTI  \cite{geiger2013kitti}.

Finally, the world coordinate system's origin (i.e. extrinsic calibration of the RGB and depth camera) can be displayed and adjusted from inside the point cloud viewer using the bounding box control scheme.

\subsection{Instance Segmentation}\label{subsec:segmentation}
The segmentation module is, in its core, aided by the GrabCut algorithm \cite{rother2004grabcut}. Initially, a user supplies a region of interest $R_k$ (represented as 2D rectangles) in the image space, containing the object that shall be segmented. This generates an initial mask $M_k$ with pixels in- and outside of $R_k$ being designated fore- and background respectively. Given $M_k$ and the corresponding image, the algorithm learns \acp{gmm} for fore- and background and improves upon the initial pixel assignment, resulting in a new mask $M'_k$, defined as:
\begin{equation}
    \label{eq:grabcut}
	M'_k = f_{GrabCut}\left(M_k\right)
\end{equation}

\begin{figure}[h]
	\centering
		\includegraphics[width=\linewidth]{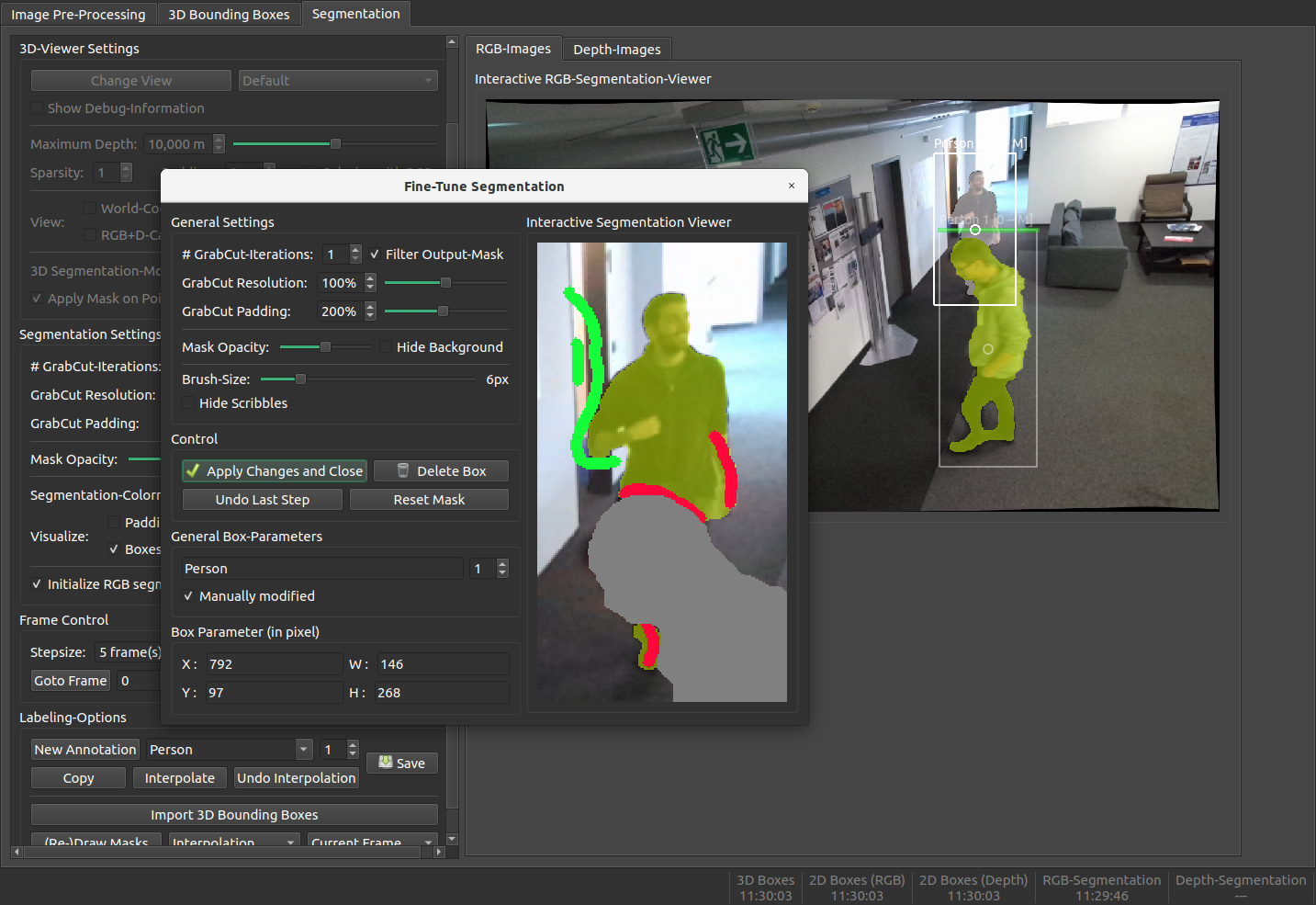}
		\caption{Screenshot of the segmentation module of SALT. Masks (yellow) are created by using GrabCut \cite{rother2004grabcut} and can be optionally fine-tuned. For this, a user provides coarse scribbles for fore- and background (red and green respectively) for the algorithm to iteratively improve upon its initial mask. Overlapping masks (gray) are used to initialize pixels as background.}
		\label{fig:segmentation}
\end{figure}

Depending on the input image, the process of Equation \ref{eq:grabcut} may not yield an optimal segmentation result. In such cases, $M'_k$ can be modified by the user in the form of coarse scribbles (cf. Figure \ref{fig:segmentation}), manually assigning fore- or background to the pixels. By running $f_{GrabCut}\left(\cdot\right)$ on the modified mask, the \acp{gmm} are updated further, improving the estimated segmentation mask. This process can be repeated iteratively until convergence. Asking the user to indicate fore- and background with coarse scribbles instead of polygons or exact pixel-level segmentation reduces the cognitive and physical workload by a wide margin.

To further improve both run time and accuracy, multiple extensions are applied to the default GrabCut implementation. First, the 2D rectangles can be created and interpolated similarly to the 3D bounding boxes, or inferred by projecting already annotated 3D bounding boxes into 2D. Furthermore, instead of generating $M_k$ by using the whole frame $F_k$, we only use the rectangle and a small, padded area outside of its boundaries. We argue that the local surroundings of objects are sufficient for the algorithm to determine background pixels. If two masks $M_{k,1}$ and $M_{k,2}$ overlap, the pixels belonging to $M_{k,2}$ can be set as background when running $f_{GrabCut}\left(M_{k,1}\right)$ (cf. Figure \ref{fig:segmentation}). Moreover, images fed into the algorithm can be downsampled beforehand to improve run time and the generated masks can be filtered using simple morphological operations. Finally, individual masks can be copied into subsequent frames, speeding up annotation time of non-moving objects.

\begin{figure}[h]
	\begin{subfigure}{0.32\linewidth}
		\includegraphics[width=\linewidth]{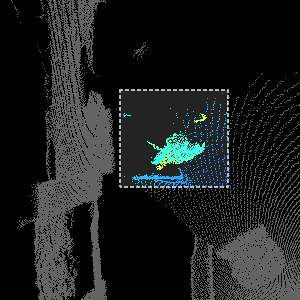}
	\end{subfigure}
	\hfill
	\begin{subfigure}{0.32\linewidth}
		\includegraphics[width=\linewidth]{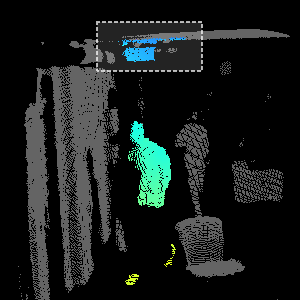}
	\end{subfigure}
	\hfill
	\begin{subfigure}{0.32\linewidth}
		\includegraphics[width=\linewidth]{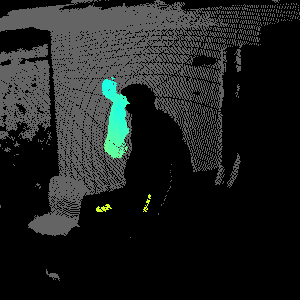}
	\end{subfigure}
	\caption{Segmenting a depth map using the 3D point cloud viewer. A user adds (left) or deletes (center) multiple points at once from different views using a 2D rectangle, resulting in a fully segmented object (right).}
	\label{fig:seg3d}
\end{figure}

Even though GrabCut relies on color cues in the RGB color space to estimate masks, we empirically found that it is also applicable to the depth frames when provided as color mapped versions. We additionally convert them into grayscale images to make it easier for the user to differentiate between scribbles, masks and actual image contents. Given the three dimensional nature of depth maps, however, it is easier to differentiate between foreground and background when looking at the scene from multiple angles. Thus, we additionally provide the option to create depth segmentation masks by using an interactive 3D point cloud viewer similar to the one used for the 3D bounding box module. The process of creating a mask in 3D is visualized in Figure \ref{fig:seg3d}.

When a new mask is created in the RGB image, the tool will optionally search for a corresponding mask in the depth map. If found, its points will be projected into the RGB image space and used to initialize pixels as foreground in $M_k$, making the initial guess as precise as possible. Depending on the quality of the camera calibration and the density of the depth map, this feature can potentially remove the need to manually fine-tune the annotation.

\section{\uppercase{Experiments}}
\label{sec:experiments}

\begin{table*}
	\centering
	\caption{Quantitative comparison of user annotations with the high-quality ground truth for both toy data sets in terms of \acf{ase} ($\downarrow$), \acf{ate} ($\downarrow$), \acf{aoe} ($\downarrow$), \acf{iou} ($\uparrow$), \acf{mae} ($\downarrow$) and $t_\text{avg}$ ($\downarrow$). Coverage ($\uparrow$) represents the percentage of annotated objects w.r.t. the ground truth (1.0 means all objects were annotated by the user). BB$_\text{3D}$ are 3D bounding box annotations, S$_\text{RGB}$ and S$_\text{depth}$ are the RGB and depth segmentation masks. Naive represents user annotations with a naive tool (averaged across all users).}
	\label{tab:comparison_user_study}
    	\begin{tabular}{c c c c c c c c c c}	
    	\toprule
    	& & \multicolumn{4}{c}{ds\_people} & \multicolumn{4}{c}{ds\_objects}\\
		\cmidrule(lr){3-6} \cmidrule{7-10}
    	Data & Metric & User 1 & User 2 & User 3 & Naive & User 1 & User 2 & User 3 & Naive\\
    	\midrule
        \multirow{5}{*}{BB$_\text{3D}$} & \acs{ase}               & 0.320 & 0.338 & 0.294 & 0.296 & 0.226 & 0.332 & 0.239 & 0.393 \\       
                                        & \acs{ate} (in cm)       & 9.781 & 9.264 & 9.182 & 8.103 & 2.543 & 1.867 & 2.253 & 2.753 \\
                                        & \acs{aoe} (in rad)      & 0.257 & 0.150 & 0.149 & 0.141 & 0.156 & 0.185 & 0.115 & 0.241 \\
                                        & $t_\text{avg}$ (in s)   & 1.53  & 1.85  & 3.86  & 51.95 & 8.62  & 7.63  & 11.93 & 88.71 \\
                                        & Coverage                & 0.960 & 0.999 & 0.966 & 1.0   & 1.0   & 1.0   & 1.0   & 1.0 \\
        \cmidrule{2-10}
        \multirow{4}{*}{S$_\text{RGB}$} & \acs{iou}               & 0.943 & 0.948 & 0.953 & 0.942 & 0.955 & 0.958 & 0.956 & 0.948 \\
                                        & \acs{mae}               & 0.015 & 0.014 & 0.012 & 0.018 & 0.011 & 0.010 & 0.011 & 0.013 \\
                                        & $t_\text{avg}$ (in s)   & 47.64 & 69.01 & 75.28 & 146.6 & 18.19 & 24.41 & 44.39 & 155.56 \\
                                        & Coverage                & 1.0   & 1.0   & 1.0   & 1.0   & 1.0   & 1.0   & 1.0   & 1.0 \\
        \cmidrule{2-10}
        \multirow{4}{*}{S$_\text{depth}$} & \acs{iou}             & 0.999 & 1.000 & 0.998 &  -    & 0.874 & 0.904 & 0.886 & - \\
                                          & \acs{mae}             & 0.000 & 0.000 & 0.000 &  -    & 0.019 & 0.014 & 0.016 & - \\
                                          & $t_\text{avg}$ (in s) & 24.40 & 31.82 & 32.87 &  -    & 42.53 & 65.55 & 55.54 & - \\
                                          & Coverage              & 0.964 & 0.976 & 0.976 &  -    & 1.0   & 0.984 & 1.0   & - \\
	    \bottomrule
    \end{tabular}
\end{table*}

\begin{figure*}
	\centering
		\begin{subfigure}{0.245\textwidth}
			\includegraphics[width=\textwidth]{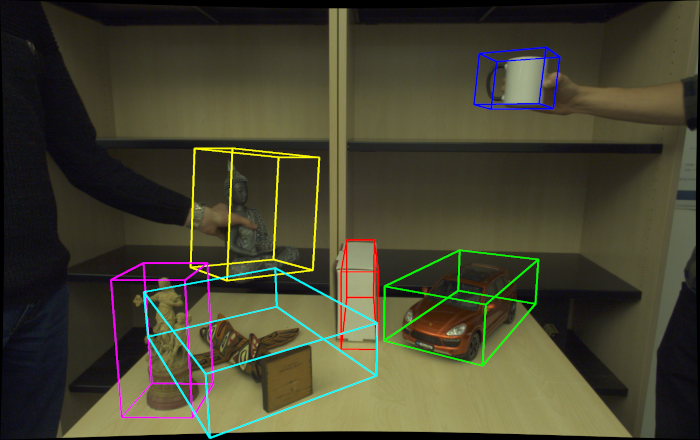}
		\end{subfigure} 
		\hfill
		\begin{subfigure}{0.245\textwidth}
			\includegraphics[width=\textwidth]{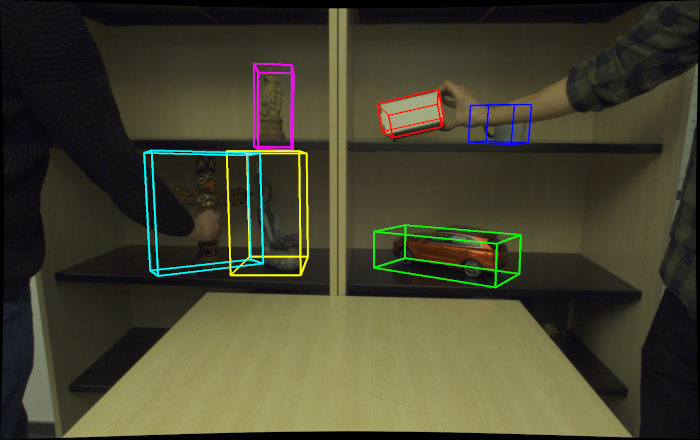}
		\end{subfigure} 
		\hfill
		\begin{subfigure}{0.245\textwidth}
			\includegraphics[width=\textwidth]{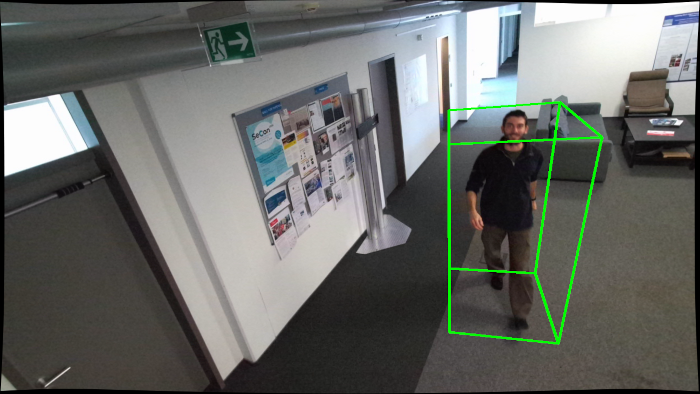}
		\end{subfigure} 
		\hfill
		\begin{subfigure}{0.245\textwidth}
			\includegraphics[width=\textwidth]{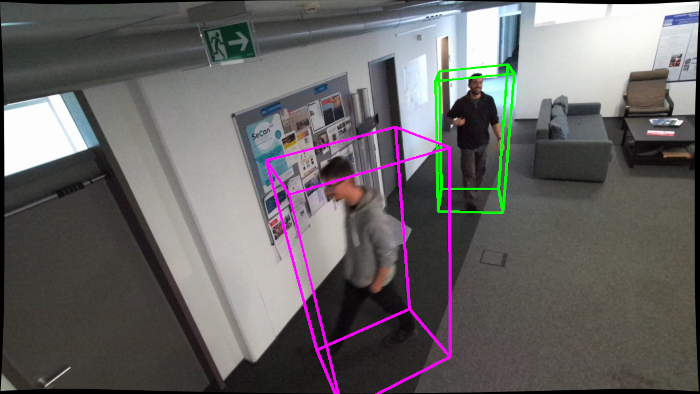}
		\end{subfigure}
		\\
		\vspace{1mm}
		\begin{subfigure}{0.245\textwidth}
			\includegraphics[width=\textwidth]{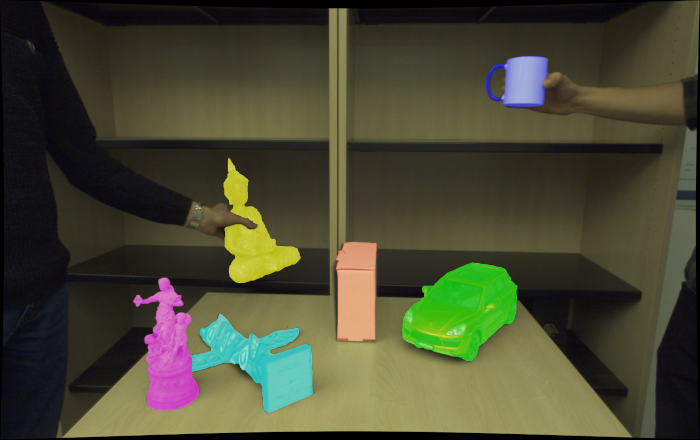}
		\end{subfigure}
		\hfill
		\begin{subfigure}{0.245\textwidth}
			\includegraphics[width=\textwidth]{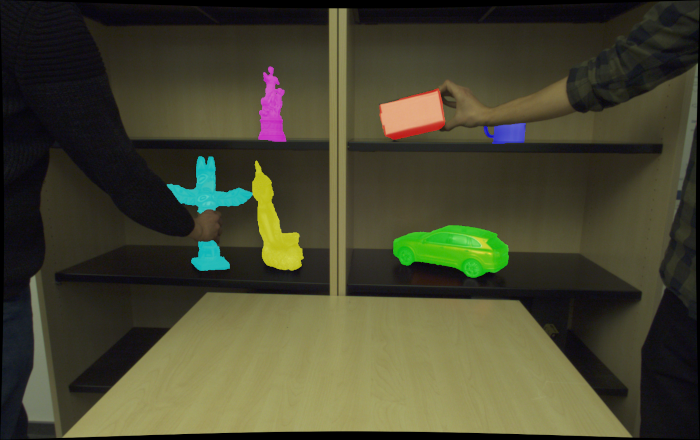}
		\end{subfigure}
		\hfill
		\begin{subfigure}{0.245\textwidth}
			\includegraphics[width=\textwidth]{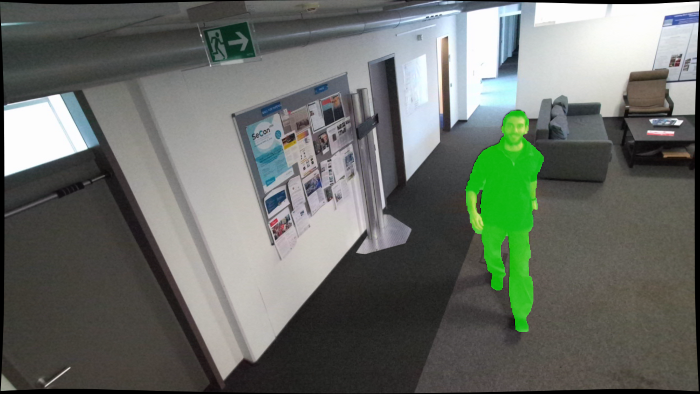}
		\end{subfigure}
		\hfill
		\begin{subfigure}{0.245\textwidth}
			\includegraphics[width=\textwidth]{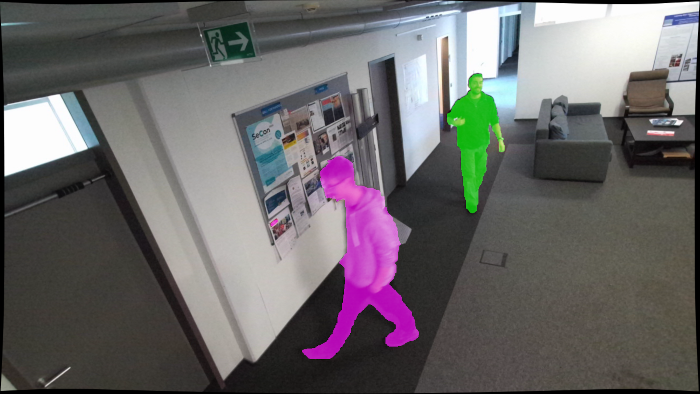}
		\end{subfigure}
		\\
		\vspace{1mm}
		\begin{subfigure}{0.245\textwidth}
			\includegraphics[width=\textwidth]{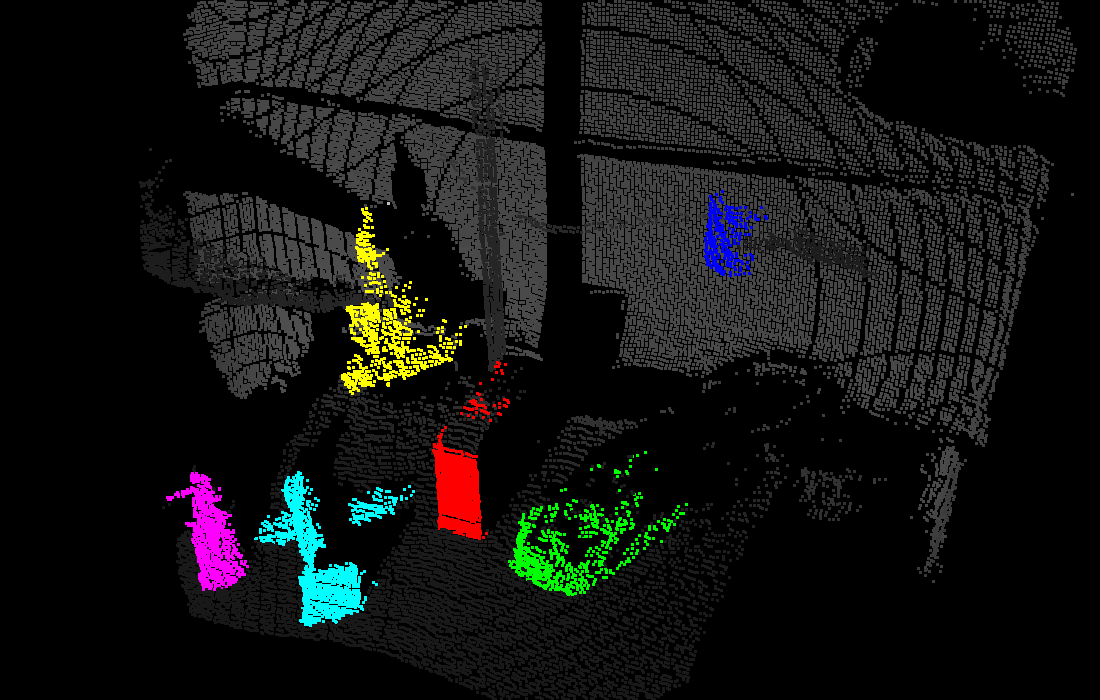}
		\end{subfigure}
		\hfill
		\begin{subfigure}{0.245\textwidth}
			\includegraphics[width=\textwidth]{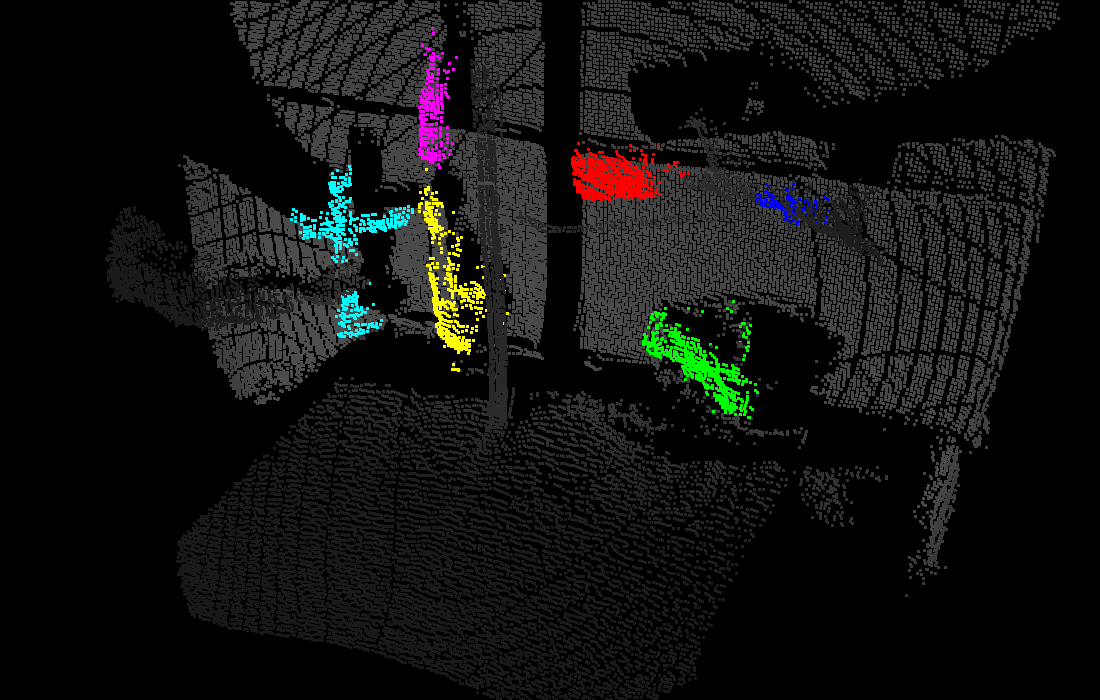}
		\end{subfigure}
		\hfill
		\begin{subfigure}{0.245\textwidth}
			\includegraphics[width=\textwidth]{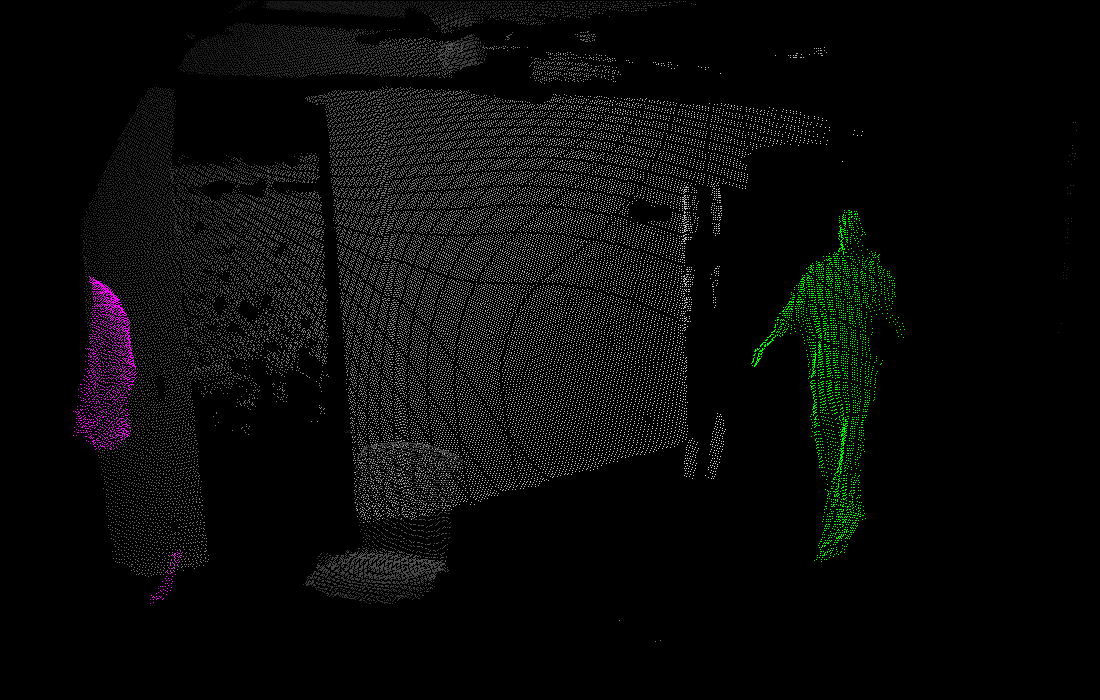}
		\end{subfigure}
		\hfill
		\begin{subfigure}{0.245\textwidth}
			\includegraphics[width=\textwidth]{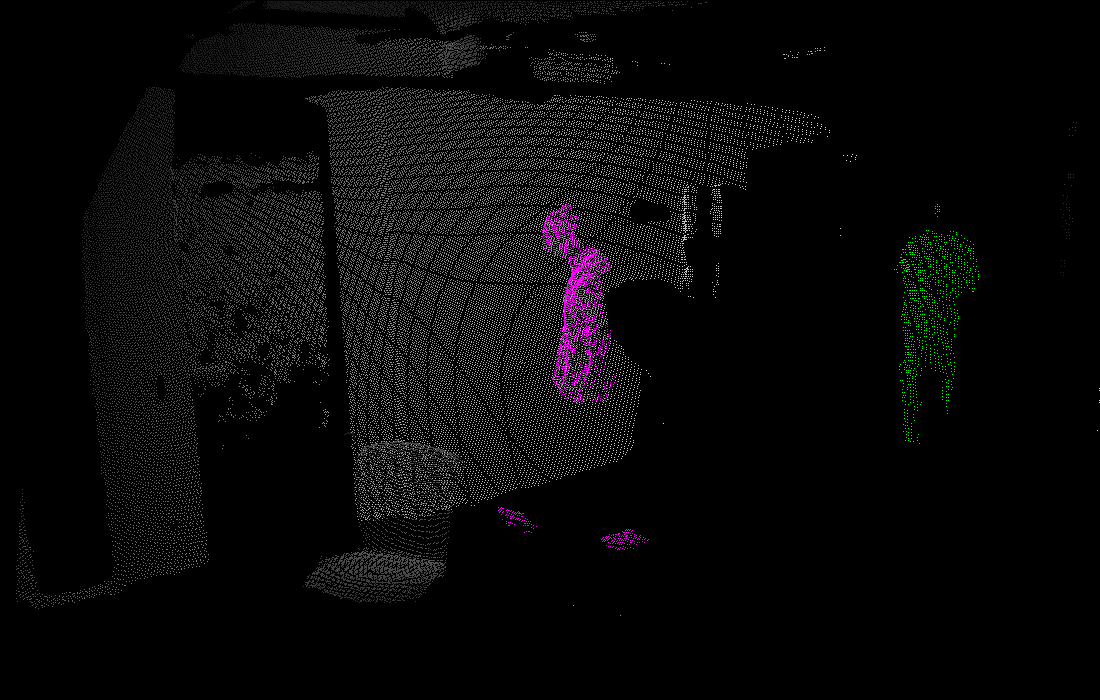}
		\end{subfigure}
	\caption{Qualitative results (3D bounding boxes, RGB and depth segmentation masks) of user annotated data using SALT. The two leftmost columns are from the ds\_objects data set, the others from ds\_people. Segmented depth maps are depicted as colorized point clouds. Due to the low depth resolution of ds\_objects, points are projected onto a 3x3 pixel region.}
	\label{fig:annotated_samples}
\end{figure*}

In order to evaluate the effectiveness of our tool, we create two toy data sets: 1000 frames of two people walking around, and 100 frames of various objects being placed into a shelf (referred to as \textit{ds\_people} and \textit{ds\_objects} respectively). The first data set can be considered easy to annotate given the high framerate and depth map quality as well as simple object poses, while reflective surfaces, low framerate, low depth map resolution and complex object poses of \textit{ds\_objects} make it a challenging data set. Annotated samples of these data sets can be viewed in Figure \ref{fig:annotated_samples}.

We asked three different users to annotate these sequences using SALT. For 3D bounding boxes, they were instructed to annotate every 40th and 5th frame of \textit{ds\_people} and \textit{ds\_objects}, respectively, as well as frames in which objects enter or leave the scene. Segmentation masks were annotated for every 20th and 5th frame, respectively. We compare the user annotations with a high-quality ground truth created with high care by a different expert user. Results are displayed in Table \ref{tab:comparison_user_study}.

When evaluating the image segmentation masks, we report the common metrics \acf{mae} and \acf{iou}:
\begin{align}
    \label{eq:eval_segmentation}
    IoU &= \frac{\sum_{j=1}^{w}\sum_{i=1}^{h}\left( M_{i,j} \cdot M_{i,j}^*\right)}{\sum_{j=1}^{w}\sum_{i=1}^{h}\left( M_{i,j} + M_{i,j}^* - M_{i,j} \cdot M_{i,j}^* \right)} \\
    MAE &= \frac{1}{h\cdot w} \sum_{j=1}^{w}\sum_{i=1}^{h} \left|M_{i,j} - M_{i,j}^* \right|\text{,}
\end{align}

\noindent
with $h$, $w$ being the height and width of the 2D rectangle area containing the mask (cf. Section \ref{subsec:segmentation}) and $M_{i,j}, M_{i,j}^* \in \{0, 1\}$ being the corresponding pixel values at position $(i, j)$ of user annotations and ground truth, respectively. 

For evaluating the 6 \ac{dof} 3D bounding boxes, we report the \acf{ase}, \acf{ate} and \acf{aoe} as introduced in the \textit{nuScenes} benchmark \cite{caesar2020nuscenes}. However, we modify the \acs{aoe} to take all three rotation angles into account, since \textit{nuScenes} only reports \textit{yaw}. More precisely, given the 3x3 rotation matrices $R_{gt}$ and $R_{u}$ for ground truth and user annotation respectively, the difference in orientation $\Delta\varphi$ can be computed as:
\begin{align}
    \label{eq:eval_aoe}
    \Delta\varphi &= arccos\left(\frac{tr\left(R_{u}R_{gt}^T\right) - 1}{2}\right)
\end{align}

All of the aforementioned evaluation metrics are averaged across all annotations of the individual data sets for each user. Furthermore, we measure the total annotation time per data set for each user, and compute the resulting average annotation time per individual object $t_\text{avg}$. For the 3D bounding boxes, both the manually created as well as the interpolated boxes are considered when computing $t_\text{avg}$.

To further evaluate the achievable annotation time speedup using SALT, we implement a naive version of the tool. This means that 3D bounding boxes are created without copying or interpolating, while segmentation masks are manually drawn without being assisted by the GrabCut algorithm. We let all three users annotate the same, randomly sampled subset of frames using the naive approach and report the average annotation time $t_\text{avg}$ and accuracy per object across all three users. As objects are not always distinguishable in the depth maps, applying this approach on the depth segmentation has empirically shown to fail. Therefore, we only report the naive results of RGB segmentation and 3D bounding boxes.

As can be seen in Table \ref{tab:comparison_user_study}, using SALT allows a reduction in annotation time without compromising quality compared to a high-quality ground truth. RGB segmentation can be sped up by a factor of up to 3.08 for ds\_people and 8.55 for ds\_objects, while 3D bounding box creation is up to 33.95 and 11.63 times faster, respectively, when compared to a naive tool. The gap between those factors is a result of the difference in difficulty for our data sets. Objects in ds\_objects only move during short periods of time, allowing masks to be copied into subsequent frames, reducing the annotation time even further. Movements in ds\_people involve less complex object poses, which makes annotating the bounding boxes easier. Even though User 3 has never used the tool before, and therefore achieves the slowest annotation time, the overall accuracy across all three users is on par and in some cases even higher than with the naive tool.

Lastly, we evaluate different interpolation approaches for 3D bounding box propagation in terms of \ac{ase} and \ac{ate}. More precisely, given the user annotated keyframes, we apply \textit{linear}, \textit{cubic} and our \textit{hybrid} interpolation (cf. Section \ref{subsec:3d_bounding_boxes}). The results are listed in Table \ref{tab:comparison_linear_cubic} and suggest that relying solely on linear interpolation is not optimal. For our data sets, we achieved best results when applying linear interpolation on the scaling parameters and our hybrid interpolation approach on the translation parameters. In case of translation, linear interpolation performs worst as it assumes constant velocity between keyframes, thus failing to capture dynamic object movements. Cubic interpolation on the other hand will induce jitter as a byproduct of the polynomial fitting shortly before movement when an object is standing still for some time. This drawback is solved by our hybrid approach, as we apply cubic interpolation only during object movement, and linear otherwise.

\begin{table}
	\centering
	\caption{Quantitative comparison of different interpolation approaches for scale and translation parameters of the 3D bounding boxes.}
	\label{tab:comparison_linear_cubic}
    \begin{tabular}{c c c c c}	
\toprule
& Metric & User 1 & User 2 & User 3 \\
\midrule
\multirow{5}{*}{\rotatebox{90}{ds\_{}people}} & \acs{ase} (linear) & 0.320 & 0.338 & 0.290 \\
                                              & \acs{ase} (cubic)  & 0.331 & 0.339 & 0.299 \\
\cmidrule{2-5}
                                              & \acs{ate} (linear) & 11.740 & 11.363 & 10.921 \\
                                              & \acs{ate} (cubic)  &  9.781 &  9.264 & 9.345 \\
                                              & \acs{ate} (hybrid) &  9.781 &  9.264 & 9.182 \\
\midrule
\multirow{5}{*}{\rotatebox{90}{ds\_{}objects}} & \acs{ase} (linear) & 0.226 & 0.332 & 0.239 \\
                                               & \acs{ase} (cubic)  & 0.225 & 0.332 & 0.239 \\
\cmidrule{2-5}
                                               & \acs{ate} (linear) & 2.685 & 1.987 & 2.330 \\
                                               & \acs{ate} (cubic)  & 2.617 & 1.898 & 2.286 \\
                                               & \acs{ate} (hybrid) & 2.543 & 1.867 & 2.253 \\
\bottomrule
\end{tabular}

\end{table}

\section{\uppercase{Conclusion}}
\label{sec:conclusion}
In this paper, we introduced SALT, a tool to semi-automatically annotate RGB-D video sequences. The tool provides a pipeline for creating 6 \ac{dof} 3D bounding boxes, as well as instance segmentation masks for both RGB images and depth maps. We have shown that by making full use of the provided features, annotation time can be reduced by a factor of up to 33.95 for 3D bounding box creation and 8.55 for RGB segmentation without compromising annotation quality. In some cases, the quality of automatically generated data even improved, demonstrating that our provided functionalities reduce the cognitive workload of the user and make the annotation process more intuitive.

For future work, several additions can be considered to further enhance the efficiency of our proposed pipeline. For example, distinguishing objects of interest from static background in 3D point clouds is a challenging task. By adding ground-plane and background removal algorithms as part of the pre-processing pipeline for depth maps, distracting elements can be removed from the scene, allowing the user to identify and annotate objects with greater ease. Furthermore, depth points inside 3D bounding boxes can be projected to an already segmented RGB image to automatically infer depth segmentation masks, leaving the user to only remove or add individual points, if necessary.

\section*{\uppercase{Acknowledgments}}
We would like to thank every participant in the user study for their time and efforts. This work was partially funded by the German Federal Ministry of Education and Research in the context of the project ENNOS (13N14975).

\bibliographystyle{apalike}
{\small\bibliography{./content/bib}}

\begin{acronym}
\acro{cnn}[CNN]{Convolutional Neural Network}
\acro{crf}[CRF]{Conditional Random Field}
\acro{dof}[DoF]{Degrees of Freedom}
\acro{slerp}[SLERP]{Spherical Linear Interpolation}
\acro{gui}[GUI]{Graphical User Interface}
\acro{gmm}[GMM]{Gaussian Mixture Model}
\acro{ase}[ASE]{Average Scaling Error}
\acro{ate}[ATE]{Average Translation Error}
\acro{aoe}[AOE]{Average Orientation Error}
\acro{mae}[MAE]{Mean Absolute Error}
\acro{iou}[IoU]{Intersection over Union}
\end{acronym}

\end{document}